\documentclass[10pt,twocolumn,letterpaper]{article}

\usepackage[pagenumbers]{iccv} 

\definecolor{iccvblue}{rgb}{0.21,0.49,0.74}
\usepackage[pagebackref,breaklinks,colorlinks,allcolors=iccvblue]{hyperref}

\usepackage{amssymb}
\usepackage{pifont}
\usepackage{utfsym}
\usepackage{multirow}
\usepackage{colortbl}
\usepackage{xcolor} 
\usepackage{amsmath}

\definecolor{light-gray}{gray}{0.85}

\def\paperID{10884} 
\def\confName{ICCV}
\def\confYear{2025}

\title{Context-Aware Academic Emotion Dataset and Benchmark}

\author{Luming Zhao{\textsuperscript{1}\thanks{Equal Contribution}} ,\, Jingwen Xuan\textsuperscript{1}\footnotemark[1] ,\, Jiamin Lou\textsuperscript{2},\, Yonghui Yu\textsuperscript{1},\, Wenwu Yang\textsuperscript{1}\thanks{Corresponding Author (wwyang@zjgsu.edu.cn)}\\
\textsuperscript{1}Zhejiang Gongshang University \quad \textsuperscript{2}Zhejiang Yuexiu University
}

\begin{document}
\maketitle
\begin{abstract}

Academic emotion analysis plays a crucial role in evaluating students' engagement and cognitive states during the learning process. This paper addresses the challenge of automatically recognizing academic emotions through facial expressions in real-world learning environments. While significant progress has been made in facial expression recognition for basic emotions, academic emotion recognition remains underexplored, largely due to the scarcity of publicly available datasets. To bridge this gap, we introduce RAER, a novel dataset comprising approximately 2,700 video clips collected from around 140 students in diverse, natural learning contexts such as classrooms, libraries, laboratories, and dormitories, covering both classroom sessions and individual study. Each clip was annotated independently by approximately ten annotators using two distinct sets of academic emotion labels with varying granularity, enhancing annotation consistency and reliability. To our knowledge, RAER is the first dataset capturing diverse natural learning scenarios. Observing that annotators naturally consider context cues—such as whether a student is looking at a phone or reading a book—alongside facial expressions, we propose CLIP-CAER (CLIP-based Context-aware Academic Emotion Recognition). Our method utilizes learnable text prompts within the vision-language model CLIP to effectively integrate facial expression and context cues from videos. Experimental results demonstrate that CLIP-CAER substantially outperforms state-of-the-art video-based facial expression recognition methods, which are primarily designed for basic emotions, emphasizing the crucial role of context in accurately recognizing academic emotions. Project page: \href{https://zgsfer.github.io/CAER}{https://zgsfer.github.io/CAER}.

\end{abstract}    
\section{Introduction}
\label{sec:intro}
Academic emotions play a crucial role in the learning process, as they are directly linked to factors such as motivation, controllability, and cognition, all of which greatly affect how effectively a student learns~\cite{ClassroomAffect2007,EmoEffect-IS2012}. Thus, accurately identifying a learner's emotional state helps in analyzing their engagement levels and understanding their learning process~\cite{Companion-Kort2001}, thereby enabling actions that positively impact learning, which not only leads to better academic outcomes but also fosters a more supportive and adaptive learning environment tailored to individual needs.

While academic emotions are internal psychological responses of learners, expert teachers excel at recognizing these emotional states by observing facial expressions and context cues, allowing them to adjust their teaching strategies accordingly. This makes it natural to develop approaches capable of automatically recognizing students' academic emotions or engagement levels from their facial expressions~\cite{HBCU-TAC2014,FEROnline-IET2019}. 
Most existing facial expression recognition (FER) methods focus on the automatic recognition of basic emotions, such as surprise, fear, disgust, happiness, sadness, anger, and neutral~\cite{DeepFERSurvey_TAC2022,AU_PIEEE2023,LANet_ICCV2023,EAC_ECCV2022}. 
Therefore, many academic emotion recognition studies based on facial expressions \cite{AffectTutor-2010,FEREvaluation-ICEIT2020,OnlineEducation-2024} have attempted to first use existing FER methods to recognize students' basic emotions and then map them to academic emotions.
However, there are significant semantic differences between  basic emotions and academic emotions~\cite{Companion-Kort2001,ControlValue-2006}, and the correspondence between them can be ambiguous. For instance, a basic expression of happiness might indicate that a student is enjoying the learning experience, or it could mean that the student is distracted by other non-academic, amusing things.

\begin{figure*}
    \centering
    \includegraphics[width=1.0\linewidth]{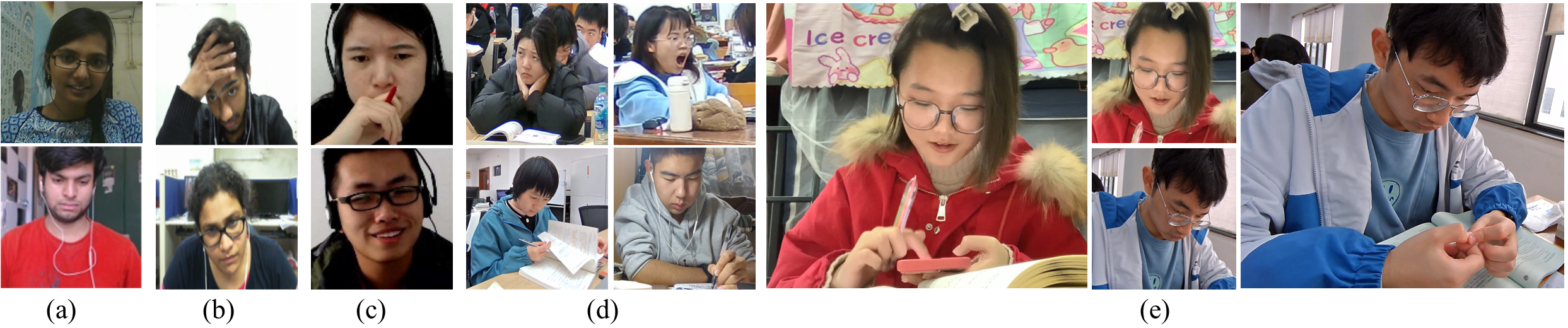}
    \vspace{-5mm}
    \caption{Examples in the (a) DAiSEE~\cite{Daisee-Arxiv2022}, (b) EngageWild~\cite{EngagementWild-DICTA2018}, (c) OL-SFED~\cite{FEROnline-IET2019}, and (d) RAER. Unlike existing datasets~\cite{Daisee-Arxiv2022,EngagementWild-DICTA2018,FEROnline-IET2019} that include emotions induced by watching stimulus videos, RAER captures natural emotions in real learning environments such as classrooms, libraries, labs, and dormitories. Moreover, RAER provides richer context information of the learning environment, which is essential for accurately identifying learners' emotions. As shown in (e), if we only consider the cropped face or upper body video (middle), it may appear that both students are engaged in studying. However, when the full context is included (left or right)—such as the female student using her phone or the male student fiddling with his fingers—it becomes clear that they are actually distracted during the learning process.
 }
    \label{fig:demo}
    \vspace{-5mm}
\end{figure*}

In this work, we investigate the automatic recognition of academic emotions from videos of students in real-world learning settings by leveraging their facial expressions. One major challenge we face is the extreme scarcity of datasets related to academic emotions in natural learning contexts. 
The number of publicly available academic emotion datasets is limited~\cite{SLR-Survey2023,SLR-Survey2023-2}, with the most commonly used ones being DAiSEE~\cite{Daisee-Arxiv2022}, EngageWild~\cite{EngagementWild-DICTA2018}, and OL-SFED~\cite{FEROnline-IET2019}. However, as shown in Fig.~\ref{fig:demo}, these datasets have several limitations: (1) They primarily focus on online learning settings, where students interact with a screen, but lack diverse natural learning scenarios, such as classroom instruction and individual study sessions in real-world environments; (2) Affective behaviors are induced by stimulus videos rather than emerging from everyday natural learning processes; (3) 
The videos capture only the learner's face or upper body, omitting essential learning contexts, which is crucial for a comprehensive representation of emotional responses. 
These limitations make it challenging for methods developed using these datasets to be effectively applied across diverse real-world learning environments.

To address these limitations, we introduce RAER, a novel dataset for Real-world Academic Emotion Recognition.
The RAER dataset comprises approximately 2,700 video clips from around 140 students, capturing diverse natural learning environments such as classrooms, libraries, laboratories, and dormitories. It includes both classroom sessions and individual study scenarios in real-world settings, as illustrated in Fig.~\ref{fig:demo}d. 
To enhance the consistency and reliability of annotations, we employ two sets of academic emotion labels with varying levels of granularity: a coarse-grained set (engaged or distracted) and a fine-grained set (enjoyment, neutrality, confusion, fatigue, or distraction). 
Specifically, our trained annotators label the learning videos using categories from a given label set, with each video independently annotated multiple times, yielding an average of approximately five annotations per label set in our experiment.
Using a majority voting strategy and cross-validating annotation consistency between the coarse- and fine-grained label sets, each video is reliably assigned a category from the fine-grained set.
To the best of our knowledge, RAER is the first real-world academic emotion dataset covering diverse natural learning contexts.

Compared to existing academic emotion datasets~\cite{Daisee-Arxiv2022, FEROnline-IET2019, EngagementWild-DICTA2018}, the RAER dataset contains videos with richer context information about the learning environment. Recognizing learners' emotions accurately often requires considering both facial expressions and context cues (see Fig.~\ref{fig:demo}e), a practice commonly employed by expert teachers and our annotators.
Building on this insight, we propose a novel framework called CLIP-based Context-aware Academic Emotion Recognition (CLIP-CAER). 
Our key idea is to leverage learnable text prompts to effectively integrate relevant facial expressions and context information from videos using the vision-language model CLIP~\cite{CLIP-ICML2021}. This approach reduces the need for extensive training data while enhancing the accuracy of academic emotion recognition.

Extensive experiments validate the effectiveness of the RAER dataset and the proposed CLIP-CAER framework. Notably, results indicate that incorporating context information significantly improves the model's ability to distinguish between distracted and engaged states during learning activities, achieving, for instance, a \textbf{19\%} accuracy increase for the ``distraction'' category in the fine-grained labels.


\subsection{Our Contributions}

\begin{itemize}
    \item We introduce the first academic emotion dataset RAER, covering a wide range of natural learning contexts. 
    \item  We propose CLIP-CAER, a novel academic emotion recognition framework that effectively integrates facial expressions with context cues in the video sequences.
    \item We demonstrate that in natural learning contexts, our method significantly outperforms state-of-the-art video-based FER approaches, highlighting the essential role of context in accurately recognizing academic emotions.
\end{itemize}

\section{Academic Emotion Database: RAER}
\label{sec:database}

\begin{figure*}
    \centering
    \includegraphics[width=1.0\linewidth]{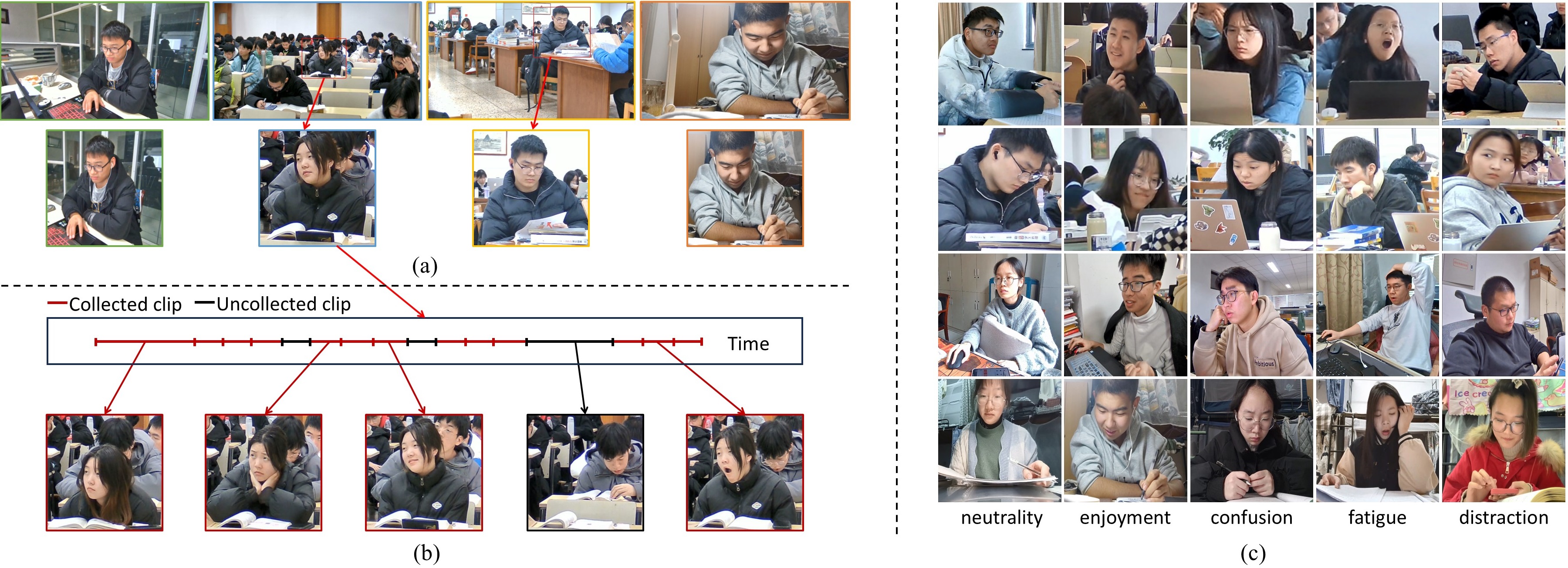}
    \vspace{-5mm}
    \caption{Procedure for building the RAER dataset: (a) extract student clips from the original videos, (b) divide each student’s clip into multiple segments, removing redundant parts so that each remaining segmented clip corresponds to an emotion. Finally, the emotion categories are annotated, with examples of the five-class academic emotions shown in (c), corresponding from top to bottom to the learning environments of the classroom, library, laboratory, and dormitory.
 }
    \label{fig:data_process}
    \vspace{-5mm}
\end{figure*}

\subsection{Collecting Learning Videos in Real World}
\label{sec:collect_video}
Unlike existing academic emotion datasets that induce academic emotions by having participants watch given stimulus videos~\cite{Daisee-Arxiv2022,EngagementWild-DICTA2018,FEROnline-IET2019} or play cognitive skills training games~\cite{HBCU-TAC2014}, we collect data on students' emotions in real-world learning settings. To achieve this goal, we contacted seven undergraduate classes at universities and additionally 
recruited 80 individuals, including both undergraduate and graduate students. The students are all adults, aged between 18 and 25.
With the consent of the teachers and students from these classes, as well as the recruited individuals, that the recorded study videos would be made publicly available solely for research purposes, we used industrial cameras to record 9 full classroom teaching sessions (covering subjects such as Mathematics, English, and Computer programming) and 177 individual study videos filmed in locations such as libraries, labs, or dormitories. During the recording of individual study sessions, the students were not subject to any special restrictions and could behave as they normally would during self-study.  For example, they were free to choose their study materials and could leave or return at their discretion.
These videos range from approximately 30 to 150 minutes in length, featuring around 332 students. 

We processed the collected videos as follows. First, for each original video, we used commercial video editing software to extract clips of each student's study session. These clips included the student's head, visible parts of their body, and the surrounding learning environment, as shown in Fig.~\ref{fig:data_process}a. During this process, we excluded students whose heads were heavily obstructed, those with unclear images due to distance, and individuals who inadvertently appeared in the videos without their knowledge.
Next, we removed redundant segments from each student's study clip that were unrelated to learning, such as when the student left their seat. Two expert teachers then reviewed each study clip and divided it into multiple segments based on noticeable emotional changes, with each segmented clip representing an academic emotion, as shown in Fig.~\ref{fig:data_process}b. For each clip, we retained only the middle 5-15 seconds, trimming any redundant content from the beginning and end.

\subsection{Coarse and Fine-grained Annotation}
\label{sec:label_video}
To annotate the video clips with emotion categories, we first need to determine the labels that will represent learners' academic emotions. Unlike basic human emotions, academic emotions are more complex, and there is currently no standardized set of labels. For instance, ~\cite{HBCU-TAC2014,EngagementWild-DICTA2018} use four engagement levels to label learning videos, ~\cite{Daisee-Arxiv2022,FEROnline-IET2019,BNU-SP2017} categorize academic emotions into 4, 5, or up to 10 types, while~\cite{RFAU-TAC2022} applies 12 action units to capture emotional nuances. 
In our dataset, we used academic emotion categories to annotate the learning videos. Building on the observation that skilled, attentive tutors and expert teachers typically respond to students by focusing on just a few common emotional cues~\cite{Companion-Kort2001}, we defined two sets of academic emotion labels with varying levels of granularity: a coarse-grained set, including engaged and distracted, and a fine-grained set, consisting of enjoyment, neutrality, confusion, fatigue, and distraction.
The coarse-grained labels offer a broad assessment of a student's state during learning activities, specifically whether they are engaged or distracted. The fine-grained labels further refine the ``engaged'' category by introducing specific emotional variations, such as enjoyment, neutrality, confusion, and fatigue. It's worth noting that we treat the ``distracted'' category as a more homogeneous condition from the standpoint of learning efficiency and, for the sake of simplicity, do not break it down further.

\begin{table*}
    \centering
    \caption{ Comparison of the RAER with existing academic emotion recognition datasets. 
    RAER provides a broader range of natural learning scenarios and includes more comprehensive  learning contexts that are essential for accurately identifying learners' emotions.}
    \vspace{-2mm}
    \label{tab:dataset_compare}
    \setlength{\tabcolsep}{4mm}{
    \begin{tabular}{lccccc}
    \hline
       Dataset  & \#Videos & \#Subjects &  Setting  &  Annotation Type & Context \\
       \hline
       \hline
DAiSEE~\cite{Daisee-Arxiv2022} & 9,068 & 112 & Online & 4 Categories & $\usym{2718}$  \\
EngageWild~\cite{EngagementWild-DICTA2018} &  195 & 78 & Online & 4 Engagement levels &  $\usym{2718}$ \\
OL-SFED~\cite{FEROnline-IET2019} & 1,274 & 82 & Online & 5 Categories & $\usym{2718}$  \\
HBCU~\cite{HBCU-TAC2014} & 120 & 34 & Controlled lab & 4 Engagement levels & $\usym{2718}$ \\
BNU-LSVED2.0~\cite{BNU-SP2017} & 2,117 & 81 & Real classrooms & 10 Categories & $\usym{2714}$  \\
RFAU~\cite{RFAU-TAC2022} & 3,325 & 1,796 & Real classrooms & 12 AUs &$\usym{2718}$ \\
\cellcolor{light-gray} & \cellcolor{light-gray} & \cellcolor{light-gray}  & \cellcolor{light-gray}Real classrooms, & \cellcolor{light-gray} & \cellcolor{light-gray} \\
\cellcolor{light-gray}  & \cellcolor{light-gray} & \cellcolor{light-gray}  & \cellcolor{light-gray} libraries, labs, & \cellcolor{light-gray} & \cellcolor{light-gray} \\
\multirow{-3}{*}{\cellcolor{light-gray}RAER} & \multirow{-3}{*}{\cellcolor{light-gray}2,649} & \multirow{-3}{*}{\cellcolor{light-gray}136} &\cellcolor{light-gray}and dormitories  & \multirow{-3}{*}{\cellcolor{light-gray}5 Categories} & \multirow{-3}{*}{\cellcolor{light-gray}$\usym{2714}$}\\
\hline
    \end{tabular} 
    }
    \vspace{-5mm}
\end{table*}

We recruited 29 annotators (students and staff from universities) to assign emotion labels to video clips. For both the coarse-grained and fine-grained label sets, each clip was assigned to the most apparent category from either the two-class or five-class set.
We developed a website for RAER annotation that displays each video with category options from a given label set. Clips were randomly and equally distributed among annotators.
Each video was independently labeled by approximately 10 annotators, with 5 assigned to the coarse-grained set and 5 to the fine-grained set.
The classification for each video in both sets was determined using a majority voting strategy. Based on the relationship between the coarse- and fine-grained labels—where enjoyment, neutrality, confusion, and fatigue correspond to the ``engaged'', and distraction is equivalent to the ``distracted''—we evaluate the consistency of each clip's classification across both label sets. The result shows that 93.5\% of the videos had consistent labels, which supports the overall reliability of the annotations.
Only videos with consistent labels were retained in the dataset. Finally, we obtained 2,649 real-world academic emotion videos, featuring 136 students (91 male and 45 female). The Fleiss' $\kappa$ coefficient~\cite{Fleiss1971} for all fine-grained labels assigned by annotators to these  videos is $0.832$, indicating a high level of consistency and reliability in the annotations.
Since the coarse-grained labels for these clips can be inferred from their fine-grained labels, each video was ultimately assigned a single category from the fine-grained set. Fig.~\ref{fig:data_process}c presents specific examples of 5-class academic emotions.


\subsection{Analysis}

Similar to ``in the wild'' expression datasets that focus on basic emotions~\cite{RAFDB-CVPR2017}, 
the academic emotions captured in RAER's natural learning scenarios also exhibit an imbalanced distribution, with neutrality accounting for 65.23\%, followed by distraction at 19.89\%, fatigue at 7.40\%, enjoyment at 6.08\%, and confusion at 1.40\%.  This imbalance highlights the differences in the distribution of academic emotions between those observed in natural learning environments and those induced by stimulus videos, where the distribution of various academic emotions tends to be more uniform, as seen in existing datasets~\cite{Daisee-Arxiv2022,EngagementWild-DICTA2018,FEROnline-IET2019,HBCU-TAC2014}.
This imbalance also aligns with our intuition to some extent, as neutrality often represents a stable emotional state during learning, while emotions such as enjoyment and confusion tend to occur less frequently, especially in structured learning environments like classrooms.

Furthermore, we compare RAER with other academic emotion datasets, such as DAiSEE~\cite{Daisee-Arxiv2022}, EngageWild~\cite{EngagementWild-DICTA2018}, OL-SFED~\cite{FEROnline-IET2019}, HBCU~\cite{HBCU-TAC2014}, BNU-LSVED2.0~\cite{BNU-SP2017}, and RFAU~\cite{RFAU-TAC2022}, as shown in Table~\ref{tab:dataset_compare}. 
DAiSEE~\cite{Daisee-Arxiv2022}, EngageWild~\cite{EngagementWild-DICTA2018}, and OL-SFED~\cite{FEROnline-IET2019} are relatively common publicly available datasets. However, these datasets primarily focus on online learning environments, where academic emotions are induced by watching stimulus videos. Additionally, they consist solely of face-cropped or upper-body-cropped videos, lacking the surrounding learning context. Furthermore,
compared to BNU-LSVED2.0~\cite{BNU-SP2017} and RFAU~\cite{RFAU-TAC2022}, which also collect videos from real classrooms, RAER includes a broader spectrum of natural learning environments, including classrooms, libraries, laboratories, and dormitories, while covering both classroom sessions and individual study. Moreover, different from RFAU~\cite{RFAU-TAC2022}, which only includes facial information, RAER provides more comprehensive context information of the learning environment. Additionally, while RFAU~\cite{RFAU-TAC2022} uses 12 action units (AUs) and 6-level intensities for each action unit in its annotations, it remains unclear how these AUs are mapped to specific academic emotion categories.



\section{Context-Aware Academic Emotion Analysis}
\label{sec:method}

\begin{figure*}
    \centering
    \includegraphics[width=0.95\linewidth]{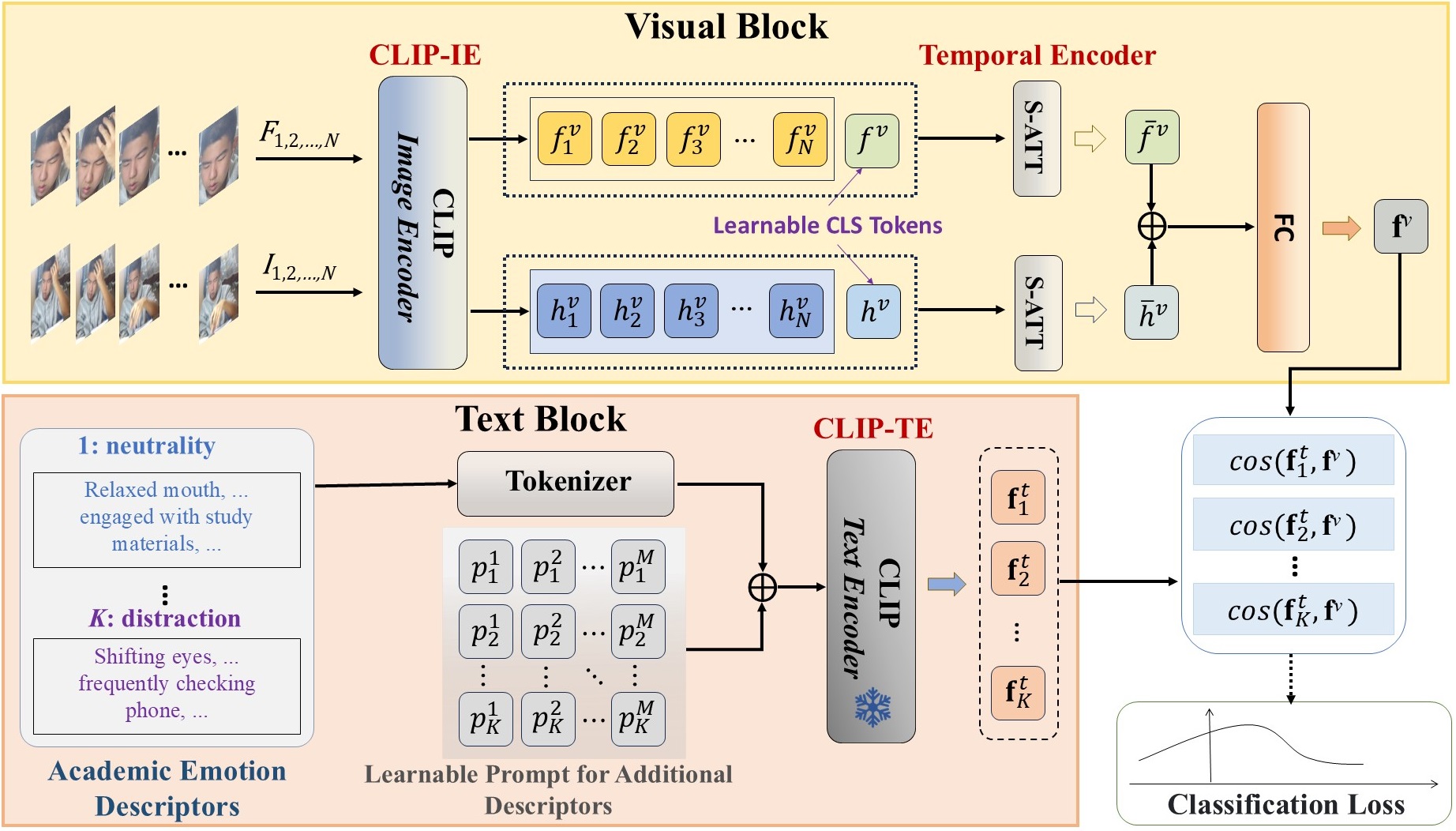}
    \caption{The framework of the proposed CLIP-CAER. Given a video sequence, we first crop the face region from each frame, forming a sequence of face images. Both the face and full-frame sequences are fed into the \textbf{Visual Block}, which outputs a visual feature token ($\textbf{f}^v$) that captures both facial expression and context information from the video. In the \textbf{Text Block}, a description along with a learnable text prompt is used to generate a text feature token for each academic emotion category (\{$\textbf{f}^t_k\}_{k=1}^K$). The input video is classified by computing the cosine similarity between the visual feature token and the text feature tokens corresponding to each academic emotion category. 
 }
    \label{fig:pipeline}
    \vspace{-5mm}
\end{figure*}


Numerous deep learning-based methods have been developed for video-based emotion recognition~\cite{DFEW-MM2020,Rethink_DFER-CVPR2023,ContextFER-ICCV2019,RNN-FER2016,LSTM-FER2019,Former-DFER2021,Transformer-FER2023,DFER-CLIP2023}. However, these methods primarily focus on recognizing basic emotions, making their adaptation to the RAER dataset for academic emotion recognition challenging for the following reasons: i) Most existing methods consider only facial expressions, neglecting context information that is crucial for accurately recognizing learners' emotions; ii) Compared to video datasets of basic emotions collected from the internet, such as DFEW~\cite{DFEW-MM2020} or MAFW~\cite{MAFW-MM2022}, the RAER dataset is significantly  smaller, making it less suitable for deep neural models like Transformers, which require large-scale training data. 
To overcome these limitations, we propose CLIP-CAER, a novel context-aware academic emotion recognition framework that harnesses the power of the vision-language pretraining model CLIP~\cite{CLIP-ICML2021}. Our approach leverages learnable text prompts within CLIP to extract and integrate relevant facial expressions and context information from videos, reducing the need for extensive training data while improving the accuracy of academic emotion recognition.

\subsection{Overview}
As shown in Fig.~\ref{fig:pipeline}, our framework, similar to the vision-language model CLIP~\cite{CLIP-ICML2021}, primarily consists of two components: a text block and a visual block. In the text block, for each academic emotion category, a fixed text is pre-generated to describe the associated facial expressions and learning contexts, complemented by a learnable text prompt to capture additional relevant details during training.
Subsequently, by inputting the fixed text and the learnable text prompt together into the \textit{CLIP text encoder}, we obtain a text feature token for each emotion category.
Given an input video, the visual block uses the \textit{CLIP image encoder} to separately extract facial expression features and context features from each video frame. These visual features are then processed through a \textit{temporal encoder module} to capture their sequential relationships, resulting in a visual feature token that effectively represents both the facial expression and context information within the video. Given the aligned visual and text feature spaces in the pre-trained CLIP model, we classify the input video by calculating the similarity between its visual feature token and the text feature tokens for each academic emotion category. 

A key distinction between previous methods that utilize the pre-trained CLIP model for recognizing basic emotions from facial expressions (\textit{e.g.}, DFER-CLIP~\cite{DFER-CLIP2023} and CLIPER~\cite{Clipper2024}) and our approach is that we integrate context cues from learning activities. This significantly enhances the model's ability to accurately identify academic emotions.
To achieve this, we design a novel context-aware temporal encoder within the visual block and incorporate context prompt into the text block, as detailed below.


\subsection{Context-Aware Visual-Text Encoding}

Given a frame sequence $\mathbf{\mathcal{S}}=\langle\mathcal{I}_1,\mathcal{I}_2,\dots,\mathcal{I}_N\rangle$ containing $N$ sample frames from an academic emotion video, the goal of the visual block is to generate a visual feature token that effectively represents both the facial expression and context information within the video, denoted as 
\begin{equation}
    \textbf{f}^v = \text{VIS-B}(\mathbf{\mathcal{S}}),
\end{equation}
where $\text{VIS-B}(\cdot)$ denotes the visual block, and $\textbf{f}^v$ is the visual feature token. In our implementation, $\textbf{f}^v \in \mathbb{R}^{512} $, matching the  class token output of the CLIP image encoder we used. 

\textbf{Context-Aware Temporal Visual Encoder.} In $\text{VIS-B}(\cdot)$, we first use a shared CLIP image encoder to extract visual features of both facial expressions and context information from each frame 
$\mathcal{I}_i$, $i \in \{1,2,\cdots,N\}$. A straightforward approach would be to directly feed each image frame into the CLIP image encoder. However, this method has two limitations: (1) Because the face region occupies a relatively small portion of the frame compared to the surrounding learning environment, the model tends to focus more on the context, potentially overlooking facial expressions; (2) The facial expression and context information for each frame are represented within the same feature vector, making it impossible to separately model their temporal dynamics. 
To address these issues, we detect and crop the face region from each frame 
$\mathcal{I}_i$
to obtain a face image $\mathcal{F}_i$. Then, we separately input $\mathcal{I}_i$ and $\mathcal{F}_i$
into the CLIP image encoder to extract context features from 
$\mathcal{I}_i$ and facial expression features from $\mathcal{F}_i$, denoted as
\begin{equation}
    \left\{
\begin{array}{c}
{f}^v_i = \text{CLIP-IE}(\mathcal{F}_i) \\
{h}^v_i = \text{CLIP-IE}(\mathcal{I}_i)
\end{array}
\right.,
\end{equation}
where $\text{CLIP-IE}(\cdot)$ denotes the CLIP image encoder, and ${{f}}_i^v$ and ${{h}}_i^v$ 
represent the feature vectors of facial expression and context information within each frame $\mathcal{I}_i$, respectively.

To effectively combine the facial expression and context features extracted from all frames in the sequence $\mathbf{\mathcal{S}}$, we employ the Transformer’s self-attention mechanism, which has proven effective in capturing long-range dependencies within sequences~\cite{Attention_NIPS2017}. 
Specifically, we apply the self-attention module separately to the facial expression feature sequence $\bar{\mathcal{S}}=\langle {{f}}^v,
 {{f}}_1^v,{{f}}_2^v,\dots,{{f}}_N^v\rangle$ and the context feature sequence $\Tilde{\mathcal{S}}=\langle {{h}}^v, 
 {{h}}_1^v, {{h}}_2^v,\dots, {{h}}_N^v\rangle$, to capture the temporal dynamics of the facial expression and context information across video frames, respectively. Note that ${{f}}^v$ in $\bar{\mathcal{S}}$ and ${{h}}^v$ in $\Tilde{\mathcal{S}}$ are special learnable vectors for the class token. After being updated, these class tokens encode the temporal information of the facial expression and context features learned by the self-attention module, respectively, \textit{i.e.}, 
\begin{equation}
\label{eqn:self_attention}
    \left\{
\begin{array}{c}
{\bar{{f}}}^v = \text{S-ATT}(\bar{\mathcal{S}}) \\
{\bar{{h}}}^v = \text{S-ATT}(\Tilde{\mathcal{S}})
\end{array}
\right.,
\end{equation}
where $\text{S-ATT}(\cdot)$ denotes the self-attention module~\cite{Attention_NIPS2017}, and ${\bar{{f}}}^v$ and ${\bar{{h}}}^v$ represent the updated class tokens for ${{f}}^v$ and ${{h}}^v$, respectively. In our implementation, the S-ATT module adheres to the conventional Transformer encoder architecture~\cite{Attention_NIPS2017}. Finally, ${\bar{{f}}}^v$ and ${\bar{{h}}}^v$ are concatenated and fed into a fully connected layer, producing the visual feature token $\textbf{f}^v$, which represents both the facial expression and context information within the academic emotion video:
\begin{equation}
   {\bar{{f}}}^v \oplus {\bar{{h}}}^v  \xrightarrow[\text{network}]{\text{ fully connected }} \textbf{f}^v,
\end{equation}
where $\oplus$ denotes the concatenation operation.

\textbf{Context-Aware Text Encoder.} 
Following the approach in ~\cite{DFER-CLIP2023}, we use text descriptions instead of class names to represent each emotion category. For each academic emotion, we describe not only the associated facial expressions but also the relevant context learning behaviors. In our implementation, rather than manually designing these descriptions, we employ a large language model, such as ChatGPT~\cite{ChatGPT-2018}, to generate them automatically. Further details are provided in the supplementary material.
Additionally, a learnable text prompt~\cite{CoOp-IJCV2022} is included to capture additional relevant details for each category during training. The structure of the prompt embeddings for each category is
\begin{equation}
    \mathcal{P}_k = [p]_k^1[p]_k^2\cdots [p]_k^M[\text{Tokenizer}( \mathcal{T}_k)],
\end{equation}
where $\mathcal{P}_k$ and $\mathcal{T}_k$ denote the prompt embeddings and the fixed text description for the $k$-th category, respectively, where $k \in \{1,2,...,K\}$, $K$ is the number of academic emotion classes, and $M$ 
is the number of tokens for the learnable text prompt, set to $M=8$ in our implementation. Each learnable token $[p]_k^j \in \mathbb{R}^{512}$, with $j \in \{ 1,2,...M \}$, matches the word embeddings input to the CLIP text encoder. Finally, each 
$\mathcal{P}_k$  is fed into the CLIP text encoder, producing the text feature token for the corresponding category:
\begin{equation}
    \textbf{f}^t_k = \text{CLIP-TE}(\mathcal{P}_k),
\end{equation}
where $\text{CLIP-TE}(\cdot)$ is the CLIP text encoder, and $\textbf{f}^t_k$ is the text feature token for the $k$-th academic emotion category. 

\textbf{Classification Loss.} 
The probability of the input video belonging to each academic emotion category is computed:
\begin{equation}
    P_{\theta}( y=k | \mathcal{S} ) = \frac{e^{\cos{(\textbf{f}^t_k, \textbf{f}^v)}}}{\sum_{j=1}^K e^{\cos{(\textbf{f}^t_j, \textbf{f}^v)}} },
\end{equation}
where $P_{\theta}(y=k| \mathcal{S} )$ represents the probability that the input video $\mathcal{S}$ belongs to the $k$-th category,  $\cos(\cdot,\cdot)$ denotes cosine similarity, and $\theta$ represents the parameters of our model, which primarily consist of a shared CLIP model and a temporal encoder module. 
We adopt the cross-entropy loss as the classification loss function. 
During training, we keep the CLIP text encoder fixed, fine-tune its image encoder, and optimize the temporal encoder module in the visual block, with the entire model trained end-to-end.

\section{Experiments}
\label{sec:experiment}

\subsection{Experimental Settings}
We have conducted evaluations on RAER for the automatic recognition of academic emotions in real-world learning scenarios. To objectively evaluate the models, we split the RAER dataset into training (80\%) and testing (20\%) sets, 
striving to exclude overlapping individuals while maintaining a nearly identical distribution of academic emotions across both sets. 
Given the class imbalance in our dataset, we use unweighted average recall (UAR), defined as the average per-class accuracy, as the evaluation metric, instead of the accuracy metric, which is sensitive to bias and ineffective for imbalanced data.
We implemented CLIP-CAER using PyTorch, utilizing the CLIP model~\cite{CLIP-ICML2021} with a ViT-B/32 architecture and its pre-trained weights. For more details, please refer to the supplementary material.

\subsection{Comparison with State-of-the-art Methods}
We first compare the performance of our proposed CLIP-CAER against current state-of-the-art video-based approaches on RAER for academic emotion recognition in natural learning contexts. Additionally, we further demonstrate the effectiveness of our method on the basic emotion dataset CAER~\cite{ContextFER-ICCV2019}, which contains context information.

\begin{figure}
    \centering
    \includegraphics[width=0.95\linewidth]{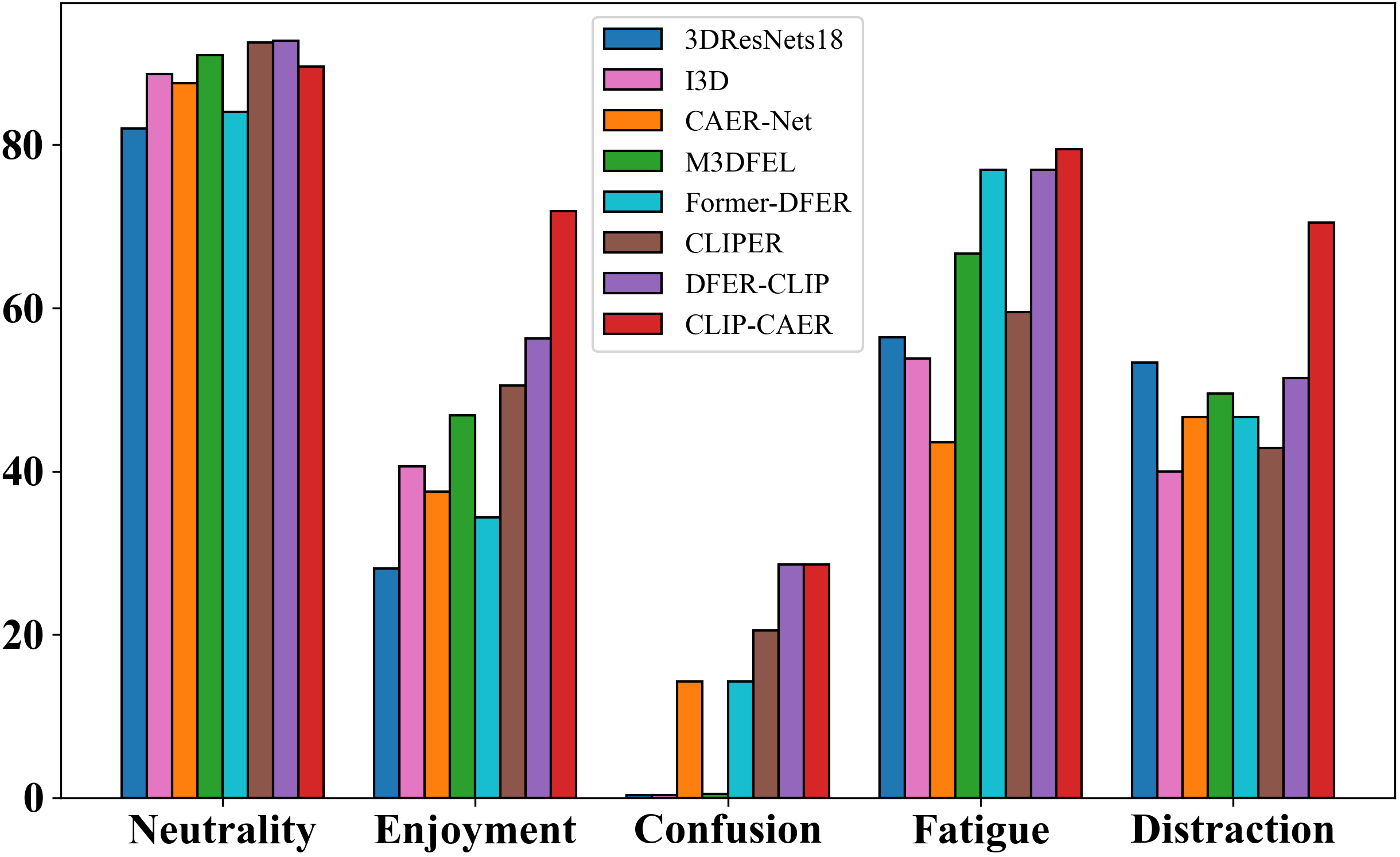}
    \vspace{-3mm}
    \caption{Comparison (UAR \%) with SOTA methods on each category in the academic emotion benchmark RAER.}
    \label{fig:class_accuracy}
    \vspace{-2mm}
\end{figure}

\begin{figure}
    \centering
    \includegraphics[width=1.0\linewidth]{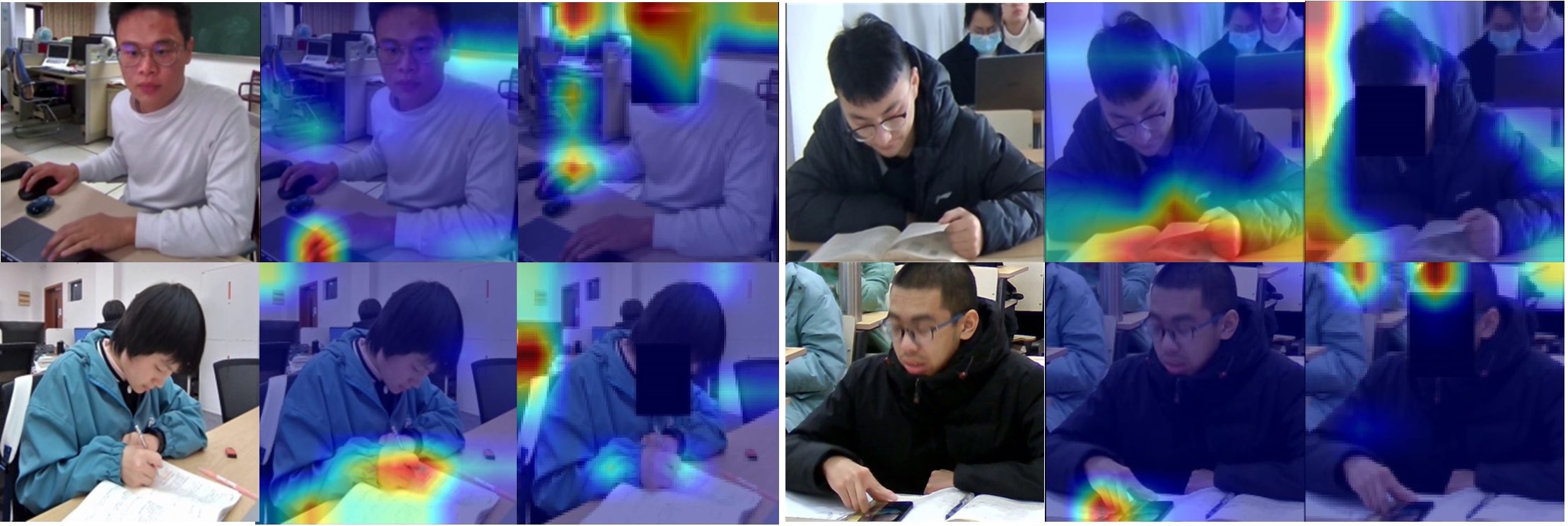}
    \vspace{-7mm}
    \caption{Visualization of context attention using Grad-CAM~\cite{GradCAM-ICCV2017}. For each example: left, input image; middle, our model; right, CAER-Net~\cite{ContextFER-ICCV2019} with face area masked as input.}
    \label{fig:attention}
    \vspace{-5mm}
\end{figure}

\textbf{Results on our RAER Dataset.} 
To ensure a fair comparison, we re-implemented several state-of-the-art methods using their publicly available code. For methods like CLIPER~\cite{Clipper2024}, DFER-CLIP~\cite{DFER-CLIP2023}, and our proposed CLIP-CAER, which are built upon the CLIP pretraining model, no additional pretraining was required. In contrast, other methods were first pre-trained on the basic emotion dataset DFEW~\cite{DFEW-MM2020} before undergoing fine-tuning on the RAER dataset. Table~\ref{tab:comparsion_RAER} presents the quantitative results of various methods on the RAER test set. The results indicate that methods using Transformers to capture temporal dynamics across frames outperform those relying on 3D CNNs for extracting video temporal information. Additionally, due to the typically smaller size of academic emotion datasets, leveraging the vision-language pre-training model CLIP has proven effective in boosting performance, as seen in the results of CLIPER~\cite{Clipper2024}, DFER-CLIP~\cite{DFER-CLIP2023}, and our CLIP-CAER. Furthermore, by integrating context information from the learning environment, our CLIP-CAER achieves the highest classification performance, outperforming the runner-up by as much as \textbf{6.81\%}. As shown in Fig.~\ref{fig:class_accuracy}, CLIP-CAER consistently outperforms across nearly all categories. Notably, it achieves a significant performance boost of nearly \textbf{20\%} in the ``distraction'' category, underscoring the critical role of context information in distinguishing between distraction and engagement. Note that although CAER-Net~\cite{ContextFER-ICCV2019} also considers context information, it struggles to capture relevant learning contexts due to the relatively small size of RAER, as demonstrated in Fig.~\ref{fig:attention}.

\begin{table}
\small
    \centering
    \caption{Comparison with SOTA methods on the academic emotion benchmark RAER. `TF' denotes Transformer.}
    \vspace{-3mm}
    \label{tab:comparsion_RAER}
    \setlength{\tabcolsep}{1.0mm}{
    \begin{tabular}{l|ccc|c|c}
        \hline
         \multirow{2}{*}{Method} &  \multicolumn{3}{c|}{Temporal Model}  &  \multirow{2}{*}{Context} & \multirow{2}{*}{UAR(\%)} \\
         & 3DCNN & LSTM & TF &  & \\
         \hline
         \hline
         3DResNets18~\cite{3DCNN-CVPR2018}                 & \checkmark &                 &                & $\usym{2718}$ & 43.98 \\
         I3D~\cite{I3D-CVPR2017}                 & \checkmark &                 &                & $\usym{2718}$ & 44.63 \\
         CAER-Net~\cite{ContextFER-ICCV2019} & \checkmark &                 &                & $\usym{2714}$ & 45.92\\
         M3DFEL~\cite{Rethink_DFER-CVPR2023} & \checkmark & \checkmark   &    \checkmark            & $\usym{2718}$ & 50.82 \\
         Former-DFER~\cite{Former-DFER2021}  &               &                 & \checkmark  & $\usym{2718}$ & 51.26 \\
         CLIPER    ~\cite{Clipper2024}      &               &    &       \checkmark         & $\usym{2718}$ & 53.18 \\
         DFER-CLIP~\cite{DFER-CLIP2023}      &               &    &      \checkmark          & $\usym{2718}$ & 61.19 \\
         \hline
         \textbf{CLIP-CAER (Ours)}                     &               &    &    \checkmark            & $\usym{2714}$ & \textbf{68.00} \\
         \hline
    \end{tabular}}
    \vspace{-5mm}
\end{table}

\textbf{Results on the CAER Dataset.}
We further evaluate our method on the basic emotion dataset CAER~\cite{ContextFER-ICCV2019}.
Due to space limitations, the results are presented in the supplementary material. These results show that the proposed CLIP-CAER, by incorporating context information, achieves an \textbf{81.77\%}  accuracy, significantly surpassing facial expression-based methods and outperforming the state-of-the-art context-aware CAER-Net~\cite{ContextFER-ICCV2019} by \textbf{4.73} points.


\subsection{Ablation Study}

\begin{figure}
    \centering
    \includegraphics[width=1.0\linewidth]{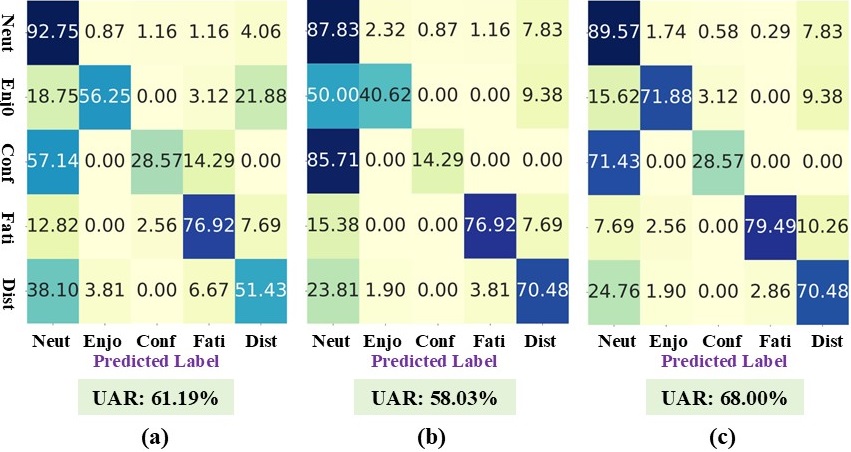}
    \vspace{-7mm}
    \caption{Ablation of different input designs in CLIP-CAER.}
    \label{fig:ablation_design}
    \vspace{-6mm}
\end{figure}

\textbf{The Design of CLIP-CAER.} Fig.~\ref{fig:ablation_design} evaluates the impact of different visual block designs: (a) using facial image sequences to model only facial expression features; (b) using full-frame video sequences to jointly capture facial expression features and context information; (c) combining facial image sequences and full-frame sequences to separately model facial expression features and context information. The results indicate that using only facial expressions yields an accuracy of 61.19\%, whereas using full-frame video sequences to jointly capture facial expressions and context information not only fails to enhance accuracy but reduces it by \textbf{3.16\%}. By combining facial image sequences with full-frame sequences, we effectively integrate facial expressions and context cues, resulting in a \textbf{6.81\%} improvement in performance. The confusion matrix in Fig.~\ref{fig:ablation_design} reveals that for the ``distraction'' category, both model (b) and model (c), which incorporate context information, achieve an accuracy of \textbf{70.48\%}, while model (a), which considers only facial expressions and ignores context information, reaches just \textbf{51.43\%}. This aligns with our observations and feedback from annotators, who often rely on context information to determine whether a learner is distracted or engaged in studying. However, with model (b), accuracy drops significantly for emotions like ``enjoyment'' and ``confusion,'' which are typically identified based on facial expressions. This supports our observation that the face region, being relatively small compared to the surrounding context, may be overlooked by this model.

\textbf{Impact of Different Prompt Strategies.}
The proposed CLIP-CAER differs from conventional classification models in that it utilizes prompts to create classifier-free predictions, making prompt engineering a crucial component of the approach. 
We compare different prompt strategies in Table~\ref{tab:ablation_prompt}.
We see that the strategy incorporating the learnable prompt consistently outperforms its counterpart without it. Moreover, our method surpasses the approach of using class names as prompts with a learnable prompt, highlighting the effectiveness of using descriptive prompts. We believe this is primarily because, compared to class names, descriptive prompts offer a more detailed and accurate representation of behaviors, including specific expressions and actions associated with academic emotions, such as yawning in fatigue, fiddling with fingers, or using a phone in distraction.


\textbf{Cross-Cultural Generalization via Indirect Validation.}
Our RAER dataset primarily includes Asian students. Unlike basic emotions, authentic academic emotions occurring in natural educational settings are difficult to obtain directly from online sources. Moreover, gathering real-world educational videos featuring students from diverse cultural backgrounds poses significant challenges, especially in a non-immigrant context. This raises a critical question: Can a model trained on RAER effectively generalize across different cultural contexts? 
To address this question, we conducted an indirect validation experiment using a newly collected dataset named JuniorRAER. Compared to RAER ($\sim$2700 videos), JuniorRAER is smaller ($<$ 470 videos). Furthermore, unlike RAER, which focuses on adult university students, JuniorRAER captures academic emotions from primary school students around 10 years old during classroom activities. Details about JuniorRAER are provided in the supplementary material.
This experiment explores the model's generalization capability beyond its original training domain, given notable differences in facial appearances, study habits, and cognitive developmental stages between primary school and university students. Additionally, variations in educational environments and cognitive engagement levels further enhance insights into the model's potential for cross-cultural applicability.

As presented in Table~\ref{tab:ablation_generalize}, the model trained on RAER demonstrates robust generalization to the JuniorRAER test set. Directly training on JuniorRAER alone leads to overfitting to the dominant neutrality class due to the dataset's limited size. Conversely, fine-tuning a model pre-trained on RAER substantially enhances performance, indicating that the RAER-trained model possesses strong generalization capabilities. We attribute the model's generalization ability to two key factors. First, 
the pre-trained CLIP model utilized in our approach was initially trained on a large-scale dataset, capturing a broad spectrum of facial features across various age groups and ethnicities. Second, fundamental academic emotions (engagement, enjoyment, confusion, fatigue, and distraction) are universally experienced across diverse educational contexts. Although specific teaching methods and cultural nuances may differ, the core emotional context associated with learning remains broadly consistent.



\begin{table}
    \centering
    \caption{Evaluation of different prompts used in CLIP-CAER.}
    \label{tab:ablation_prompt}
    \vspace{-3mm}
    \begin{tabular}{lc}
    \hline
        Prompts & UAR(\%)\\
    \hline
    \hline
         {an emotion of [Class] during studying} & 62.14 \\
         {[Learnable Prompt]}{[Class] during studying} & 64.26\\
         {[Descriptors]} & 65.43\\
         {[Learnable Prompt]} {[Descriptors]} & \textbf{68.00}\\
    \hline
    \end{tabular}
    \vspace{-1mm}
\end{table}

\begin{table}
    \centering
    \caption{Evaluation of CLIP-CAER on JuniorRAER test set: a) trained on RAER; b) trained on JuniorRAER; c) fine-tuned on a). }
    \label{tab:ablation_generalize}
    \vspace{-3mm}
    \setlength{\tabcolsep}{1.2mm}{
    \begin{tabular}{c|cccccc}
    \hline
    Models & Neut. & Enjo. & Conf. & Fati. & Dist. & UAR (\%) \\
         \hline
         \hline
         a) & 58.16 & 10.00 & 0.00 & 75.00 & 64.00 & 41.43 \\
         b) & 100.00 & 0.00 & 0.00 & 0.00 & 0.00 & 20.00 \\
         c) & 95.04 & 80.00 & 0.00 & 66.67 & 64.30 & 61.20 \\
    \hline
    \end{tabular}}
    \vspace{-5mm}
\end{table}


\section{Conclusion}
In this paper, we introduce RAER, a dataset for academic emotion recognition in real-world learning scenarios. 
Additionally, we propose CLIP-CAER, which uses CLIP with learnable text prompts to  combine facial expressions and context cues. 
Our method significantly outperforms state-of-the-art methods, 
emphasizing the crucial role of context in accurately recognizing academic emotions.
Our work provides a foundational dataset and approach for advancing research in academic emotion analysis.

\noindent\textbf{Acknowledgment:}  This work was supported by ``Pioneer'' and ``Leading
 Goose'' R\&D Program of Zhejiang Province (2024C01167) and the Fundamental Research Funds for the Provincial Universities of Zhejiang (FR24005Z).

{
    \small
    \bibliographystyle{ieeenat_fullname}
    \bibliography{main}

\begin{thebibliography}{38}
\providecommand{\natexlab}[1]{#1}
\providecommand{\url}[1]{\texttt{#1}}
\expandafter\ifx\csname urlstyle\endcsname\relax
  \providecommand{\doi}[1]{doi: #1}\else
  \providecommand{\doi}{doi: \begingroup \urlstyle{rm}\Url}\fi

\bibitem[Aly(2024)]{OnlineEducation-2024}
Mohammed Aly.
\newblock Revolutionizing online education: Advanced facial expression recognition for real-time student progress tracking via deep learning model.
\newblock \emph{Multimedia Tools and Applications}, 2024.

\bibitem[Baddar and Ro(2019)]{LSTM-FER2019}
Wissam~J. Baddar and Yong~Man Ro.
\newblock Mode variational lstm robust to unseen modes of variation: application to facial expression recognition.
\newblock In \emph{AAAI}, 2019.

\bibitem[{Ben Ammar} et~al.(2010){Ben Ammar}, Neji, Alimi, and Gouardères]{AffectTutor-2010}
Mohamed {Ben Ammar}, Mahmoud Neji, Adel.~M. Alimi, and Guy Gouardères.
\newblock The affective tutoring system.
\newblock \emph{Expert Systems with Applications}, 37\penalty0 (4):\penalty0 3013--3023, 2010.

\bibitem[Bian et~al.(2019)Bian, Zhang, Yang, Bi, and Lu]{FEROnline-IET2019}
Cunling Bian, Ya Zhang, Fei Yang, Wei Bi, and Weigang Lu.
\newblock Spontaneous facial expression database for academic emotion inference in online learning.
\newblock \emph{IET Computer Vision}, 13\penalty0 (3):\penalty0 329--337, 2019.

\bibitem[Boekaerts(2007)]{ClassroomAffect2007}
Monique Boekaerts.
\newblock Understanding students' affective processes in the classroom.
\newblock In \emph{Emotion in Education}, pages 37--56. 2007.

\bibitem[Carreira and Zisserman(2017)]{I3D-CVPR2017}
João Carreira and Andrew Zisserman.
\newblock Quo vadis, action recognition? a new model and the kinetics dataset.
\newblock In \emph{CVPR}, pages 4724--4733, 2017.

\bibitem[Fan et~al.(2016)Fan, Lu, Li, and Liu]{RNN-FER2016}
Yin Fan, Xiangju Lu, Dian Li, and Yuanliu Liu.
\newblock Video-based emotion recognition using cnn-rnn and c3d hybrid networks.
\newblock In \emph{Proceedings of the 18th ACM International Conference on Multimodal Interaction}, page 445–450, 2016.

\bibitem[Fang et~al.(2023)Fang, Li, Han, and He]{SLR-Survey2023-2}
Bei Fang, Xian Li, Guangxin Han, and Juhou He.
\newblock Facial expression recognition in educational research from the perspective of machine learning: A systematic review.
\newblock \emph{IEEE Access}, 11:\penalty0 112060--112074, 2023.

\bibitem[Fleiss(1971)]{Fleiss1971}
Joseph~L. Fleiss.
\newblock Measuring nominal scale agreement among many raters.
\newblock \emph{Psychological Bulletin}, 76\penalty0 (5):\penalty0 378–382, 1971.

\bibitem[Gupta et~al.(2022)Gupta, D'Cunha, Awasthi, and Balasubramanian]{Daisee-Arxiv2022}
Abhay Gupta, Arjun D'Cunha, Kamal Awasthi, and Vineeth Balasubramanian.
\newblock Daisee: Towards user engagement recognition in the wild, 2022.

\bibitem[Hara et~al.(2018)Hara, Kataoka, and Satoh]{3DCNN-CVPR2018}
Kensho Hara, Hirokatsu Kataoka, and Yutaka Satoh.
\newblock Can spatiotemporal 3d cnns retrace the history of 2d cnns and imagenet?
\newblock In \emph{CVPR}, pages 6546--6555, 2018.

\bibitem[Hu et~al.(2022)Hu, Mei, Jiang, Shen, Zhang, Wang, and Zhang]{RFAU-TAC2022}
Qiaoping Hu, Chuanneng Mei, Fei Jiang, Ruimin Shen, Yitian Zhang, Ce Wang, and Junpeng Zhang.
\newblock Rfau: A database for facial action unit analysis in real classrooms.
\newblock \emph{IEEE Transactions on Affective Computing}, 13\penalty0 (3):\penalty0 1452--1465, 2022.

\bibitem[Jiang et~al.(2020)Jiang, Zong, Zheng, Tang, Xia, Lu, and Liu]{DFEW-MM2020}
Xingxun Jiang, Yuan Zong, Wenming Zheng, Chuangao Tang, Wanchuang Xia, Cheng Lu, and Jiateng Liu.
\newblock Dfew: A large-scale database for recognizing dynamic facial expressions in the wild.
\newblock In \emph{ACM MM}, page 2881–2889, 2020.

\bibitem[Kaur et~al.(2018)Kaur, Mustafa, Mehta, and Dhall]{EngagementWild-DICTA2018}
Amanjot Kaur, Aamir Mustafa, Love Mehta, and Abhinav Dhall.
\newblock Prediction and localization of student engagement in the wild.
\newblock In \emph{2018 Digital Image Computing: Techniques and Applications (DICTA)}, pages 1--8, 2018.

\bibitem[Kim and Hodges(2012)]{EmoEffect-IS2012}
ChanMin Kim and Charles~B. Hodges.
\newblock Effects of an emotion control treatment on academic emotions, motivation and achievement in an online mathematics course.
\newblock \emph{Instructional Science}, 40\penalty0 (1):\penalty0 173--192, 2012.

\bibitem[Kort et~al.(2001)Kort, Reilly, and Picard]{Companion-Kort2001}
B. Kort, R. Reilly, and R.W. Picard.
\newblock An affective model of interplay between emotions and learning: reengineering educational pedagogy-building a learning companion.
\newblock In \emph{Proceedings IEEE International Conference on Advanced Learning Technologies}, pages 43--46, 2001.

\bibitem[Lee et~al.(2019)Lee, Kim, Kim, Park, and Sohn]{ContextFER-ICCV2019}
Jiyoung Lee, Seungryong Kim, Sunok Kim, Jungin Park, and Kwanghoon Sohn.
\newblock { Context-Aware Emotion Recognition Networks }.
\newblock In \emph{ICCV}, pages 10142--10151, 2019.

\bibitem[Lek and Teo(2023)]{SLR-Survey2023}
Jeniffer Xin-Ying Lek and Jason Teo.
\newblock Academic emotion classification using fer: A systematic review.
\newblock \emph{Human Behavior and Emerging Technologies}, 2023\penalty0 (1):\penalty0 9790005, 2023.

\bibitem[Li et~al.(2024)Li, Niu, Zhu, and Zhao]{Clipper2024}
Hanting Li, Hongjing Niu, Zhaoqing Zhu, and Feng Zhao.
\newblock Cliper: A unified vision-language framework for in-the-wild facial expression recognition.
\newblock In \emph{2024 IEEE International Conference on Multimedia and Expo (ICME)}, pages 1--6, 2024.

\bibitem[Li and Deng(2022)]{DeepFERSurvey_TAC2022}
Shan Li and Weihong Deng.
\newblock Deep facial expression recognition: A survey.
\newblock \emph{IEEE Transactions on Affective Computing}, 13\penalty0 (3):\penalty0 1195--1215, 2022.

\bibitem[Li et~al.(2017)Li, Deng, and Du]{RAFDB-CVPR2017}
Shan Li, Weihong Deng, and JunPing Du.
\newblock Reliable crowdsourcing and deep locality-preserving learning for expression recognition in the wild.
\newblock In \emph{CVPR}, pages 2584--2593, 2017.

\bibitem[Liu et~al.(2022)Liu, Dai, Feng, Wang, Yin, Zeng, and Shan]{MAFW-MM2022}
Yuanyuan Liu, Wei Dai, Chuanxu Feng, Wenbin Wang, Guanghao Yin, Jiabei Zeng, and Shiguang Shan.
\newblock Mafw: A large-scale, multi-modal, compound affective database for dynamic facial expression recognition in the wild.
\newblock In \emph{ACM MM}, page 24–32, 2022.

\bibitem[Liu et~al.(2023)Liu, Wang, Feng, Zhang, Chen, and Zhan]{Transformer-FER2023}
Yuanyuan Liu, Wenbin Wang, Chuanxu Feng, Haoyu Zhang, Zhe Chen, and Yibing Zhan.
\newblock Expression snippet transformer for robust video-based facial expression recognition.
\newblock \emph{Pattern Recognition}, 138:\penalty0 109368, 2023.

\bibitem[Pekrun(2006)]{ControlValue-2006}
Reinhard Pekrun.
\newblock The control-value theory of achievement emotions: Assumptions, corollaries, and implications for educational research and practice.
\newblock \emph{Educ Psychol Rev}, 18:\penalty0 315–341, 2006.

\bibitem[Radford et~al.(2018)Radford, Narasimhan, Salimans, and Sutskever]{ChatGPT-2018}
Alec Radford, Karthik Narasimhan, Tim Salimans, and Ilya Sutskever.
\newblock Improving language understanding by generative pre-training.
\newblock 2018.

\bibitem[Radford et~al.(2021)Radford, Kim, Hallacy, Ramesh, Goh, Agarwal, Sastry, Askell, Mishkin, Clark, Krueger, and Sutskever]{CLIP-ICML2021}
Alec Radford, Jong~Wook Kim, Chris Hallacy, Aditya Ramesh, Gabriel Goh, Sandhini Agarwal, Girish Sastry, Amanda Askell, Pamela Mishkin, Jack Clark, Gretchen Krueger, and Ilya Sutskever.
\newblock Learning transferable visual models from natural language supervision.
\newblock In \emph{ICML}, page 8748–8763, 2021.

\bibitem[Selvaraju et~al.(2017)Selvaraju, Cogswell, Das, Vedantam, Parikh, and Batra]{GradCAM-ICCV2017}
Ramprasaath~R. Selvaraju, Michael Cogswell, Abhishek Das, Ramakrishna Vedantam, Devi Parikh, and Dhruv Batra.
\newblock Grad-cam: Visual explanations from deep networks via gradient-based localization.
\newblock In \emph{ICCV}, pages 618--626, 2017.

\bibitem[Tang et~al.(2020)Tang, Peng, Liu, and Xiong]{FEREvaluation-ICEIT2020}
Xiao-Yu Tang, Wang-Yue Peng, Si-Rui Liu, and Jian-Wen Xiong.
\newblock Classroom teaching evaluation based on facial expression recognition.
\newblock In \emph{Proceedings of the 2020 9th International Conference on Educational and Information Technology}, page 62–67, 2020.

\bibitem[Vaswani et~al.(2017)Vaswani, Shazeer, Parmar, Uszkoreit, Jones, Gomez, Kaiser, and Polosukhin]{Attention_NIPS2017}
Ashish Vaswani, Noam Shazeer, Niki Parmar, Jakob Uszkoreit, Llion Jones, Aidan~N. Gomez, \L{}ukasz Kaiser, and Illia Polosukhin.
\newblock Attention is all you need.
\newblock In \emph{NIPS}, page 6000–6010, 2017.

\bibitem[Wang et~al.(2023)Wang, Li, Wu, Shen, Liu, Ding, and Zhou]{Rethink_DFER-CVPR2023}
Hanyang Wang, Bo Li, Shuang Wu, Siyuan Shen, Feng Liu, Shouhong Ding, and Aimin Zhou.
\newblock Rethinking the learning paradigm for dynamic facial expression recognition.
\newblock In \emph{CVPR}, pages 17958--17968, 2023.

\bibitem[Wei et~al.(2017)Wei, Sun, He, and Yu]{BNU-SP2017}
Qinglan Wei, Bo Sun, Jun He, and Lejun Yu.
\newblock Bnu-lsved 2.0: Spontaneous multimodal student affect database with multi-dimensional labels.
\newblock \emph{Signal Processing: Image Communication}, 59:\penalty0 168--181, 2017.

\bibitem[Whitehill et~al.(2014)Whitehill, Serpell, Lin, Foster, and Movellan]{HBCU-TAC2014}
Jacob Whitehill, Zewelanji Serpell, Yi-Ching Lin, Aysha Foster, and Javier~R. Movellan.
\newblock The faces of engagement: Automatic recognition of student engagementfrom facial expressions.
\newblock \emph{IEEE Transactions on Affective Computing}, 5\penalty0 (1):\penalty0 86--98, 2014.

\bibitem[Wu and Cui(2023)]{LANet_ICCV2023}
Zhi-Fan Wu and Jinshi Cui.
\newblock La-net: Landmark-aware learning for reliable facial expression recognition under label noise.
\newblock In \emph{ICCV}, pages 20698--20707, 2023.

\bibitem[Yang et~al.(2023)Yang, Hristov, Shen, Lin, and Pantic]{AU_PIEEE2023}
Jing Yang, Yordan Hristov, Jie Shen, Yiming Lin, and Maja Pantic.
\newblock Toward robust facial action units’ detection.
\newblock \emph{Proceedings of the IEEE}, 111\penalty0 (10):\penalty0 1198--1214, 2023.

\bibitem[Zhang et~al.(2022)Zhang, Wang, Ling, and Deng]{EAC_ECCV2022}
Yuhang Zhang, Chengrui Wang, Xu Ling, and Weihong Deng.
\newblock Learn from all: Erasing attention consistency for noisy label facial expression recognition.
\newblock In \emph{ECCV}, pages 418--434, 2022.

\bibitem[Zhao and Liu(2021)]{Former-DFER2021}
Zengqun Zhao and Qingshan Liu.
\newblock Former-dfer: Dynamic facial expression recognition transformer.
\newblock In \emph{ACM MM}, page 1553–1561, 2021.

\bibitem[Zhao and Patras(2023)]{DFER-CLIP2023}
Zengqun Zhao and Ioannis Patras.
\newblock Prompting visual-language models for dynamic facial expression recognition.
\newblock In \emph{BMVC}, pages 1--14, 2023.

\bibitem[Zhou et~al.(2022)Zhou, Yang, Loy, and Liu]{CoOp-IJCV2022}
Kaiyang Zhou, Jingkang Yang, Chen~Change Loy, and Ziwei Liu.
\newblock Learning to prompt for vision-language models.
\newblock \emph{International Journal of Computer Vision}, 130:\penalty0 2337–2348, 2022.

\end{thebibliography}
}

\clearpage
\setcounter{page}{1}
\maketitlesupplementary

\section*{Appendix}
\label{sec:appendix}

In the supplementary material, we provide:\\\\
\noindent{\S}\textcolor[rgb]{1.00,0.00,0.00}{A} Related Work. \\\\
\noindent{\S}\textcolor[rgb]{1.00,0.00,0.00}{B} Additional Implementation Details.\\\\
\noindent{\S}\textcolor[rgb]{1.00,0.00,0.00}{C} Experiments on CAER Dataset. \\\\
\noindent{\S}\textcolor[rgb]{1.00,0.00,0.00}{D} Additional Ablation Study.  \\\\
\noindent{\S}\textcolor[rgb]{1.00,0.00,0.00}{E} Ethical Implications. 

\section*{A. Related Work}
\label{sec:relatework}

\noindent\textbf{Academic Emotion Datasets.} 
Although there are numerous well-known publicly available datasets for basic emotions, such as RAF-DB~\cite{RAFDB-CVPR2017}, DFEW~\cite{DFEW-MM2020}, and MAFW~\cite{MAFW-MM2022}, the availability of academic emotion datasets remains limited, which significantly hinders the progress of research in academic emotion recognition.
Existing academic emotion datasets can be broadly classified into two categories: those focused on online learning environments and those focused on real-world classroom settings. For datasets focused on online learning environments, such as HBCU~\cite{HBCU-TAC2014}, 
DAiSEE~\cite{Daisee-Arxiv2022}, EngageWild~\cite{EngagementWild-DICTA2018}, and OL-SFED~\cite{FEROnline-IET2019}, participants typically interact with a computer screen while watching stimulus videos or playing cognitive skill training games to elicit academic emotions. To collect spontaneous emotions in real-world learning environments, \cite{BNU-SP2017} introduced the academic emotion dataset BNU-LSVED2.0, which contains 2,117 videos of students engaged in real classroom scenarios. Although the reliability of the BNU-LSVED2.0's annotations was assessed  using statistical methods in~\cite{BNU-SP2017}, 
experimental validation of automatic academic emotion 
recognition algorithms on this dataset is still lacking. Additionally, \cite{RFAU-TAC2022} introduced a manually annotated facial action unit (AU) database collected from juveniles in real classroom settings. However, the challenge of mapping these AUs to specific academic emotion categories remains unresolved. Overall, existing academic emotion datasets have the following limitations: 1) They lack diversity in natural learning scenarios; 2) They typically include only the learner’s face or upper body, missing the context information from the learning environment that is crucial for a comprehensive representation of emotional responses. 

\begin{figure}
    \centering
    \includegraphics[width=1.0\linewidth]{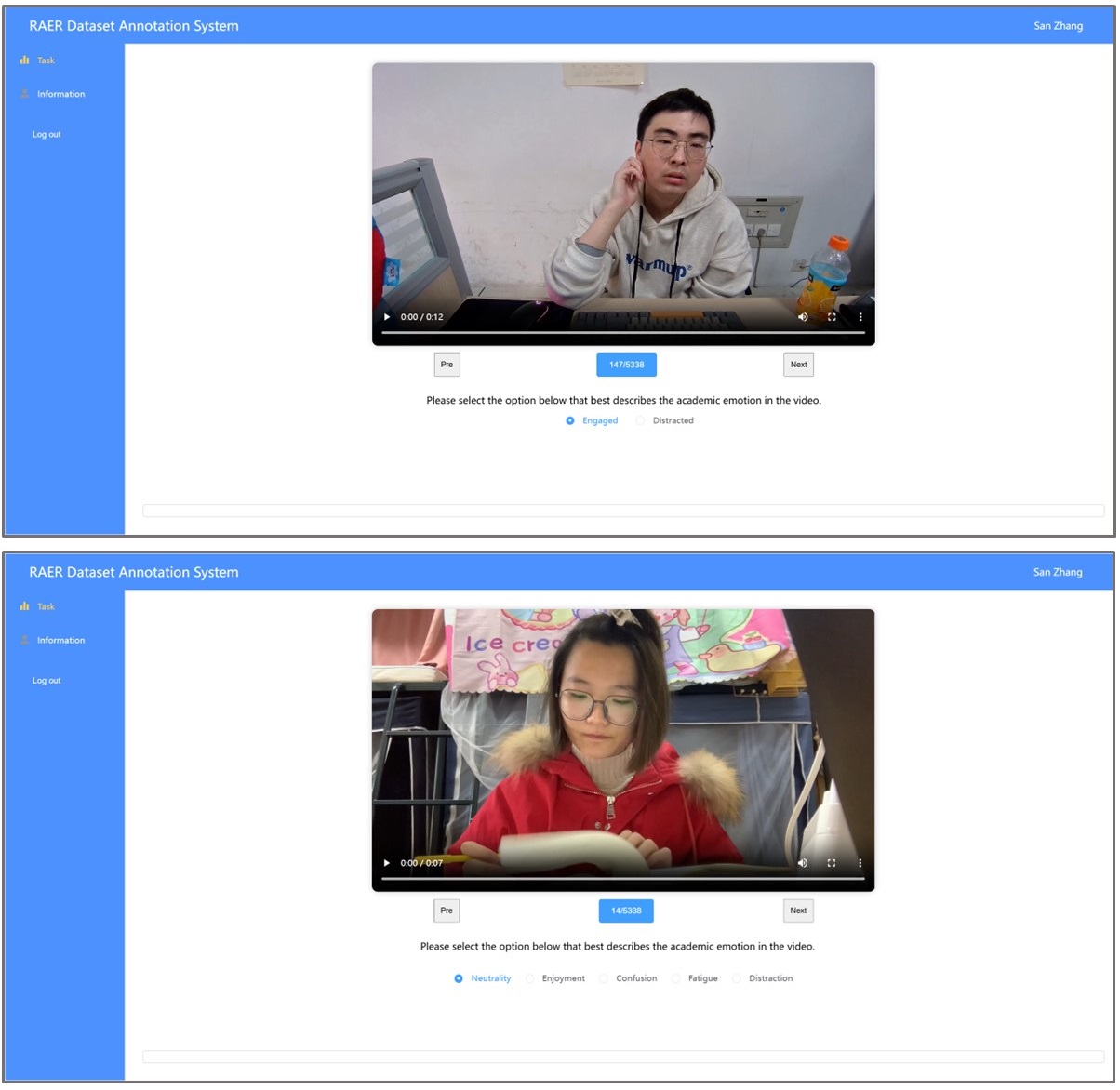}
    \caption{User interface of the annotation website developed.
 }
    \label{fig:website}
    \vspace{-8mm}
\end{figure}

In this work, we introduce the first academic emotion dataset that captures a diverse range of natural learning scenarios, including classrooms, libraries, laboratories, and dormitories, encompassing both classroom sessions and individual study in real-world settings, while providing comprehensive context information. 

\noindent\textbf{Video-Based Academic Emotion Recognition.}
Numerous deep learning-based methods have been developed for video-based emotion recognition: 3D CNN-based~\cite{DFEW-MM2020,Rethink_DFER-CVPR2023,ContextFER-ICCV2019}, RNN-based~\cite{RNN-FER2016,LSTM-FER2019}, and Transformer-based~\cite{Former-DFER2021,Transformer-FER2023,DFER-CLIP2023}. Among these, Transformer-based ones achieve state-of-the-art performance, largely due to the strength of the Transformer's attention mechanism in modeling global dependencies, allowing for more effective long-range feature extraction. However, these methods primarily focus on recognizing basic emotions, and extending them to academic emotion recognition is not  straightforward for the following reasons: i) Most existing methods consider only facial expressions, overlooking context information that is crucial for accurately recognizing learners' emotions; ii) Compared to video datasets of basic emotions collected from the internet, such as DFEW~\cite{DFEW-MM2020} or MAFW~\cite{MAFW-MM2022}, the academic emotion datasets are typically much smaller, making them less suitable for deep neural models like Transformers, which rely on large-scale training data. Therefore, it is essential to develop a specialized academic emotion recognition framework that can effectively leverage context information from various real-world learning environments without relying on large-scale training data.

\begin{figure*}
    \centering
    \includegraphics[width=1.0\linewidth]{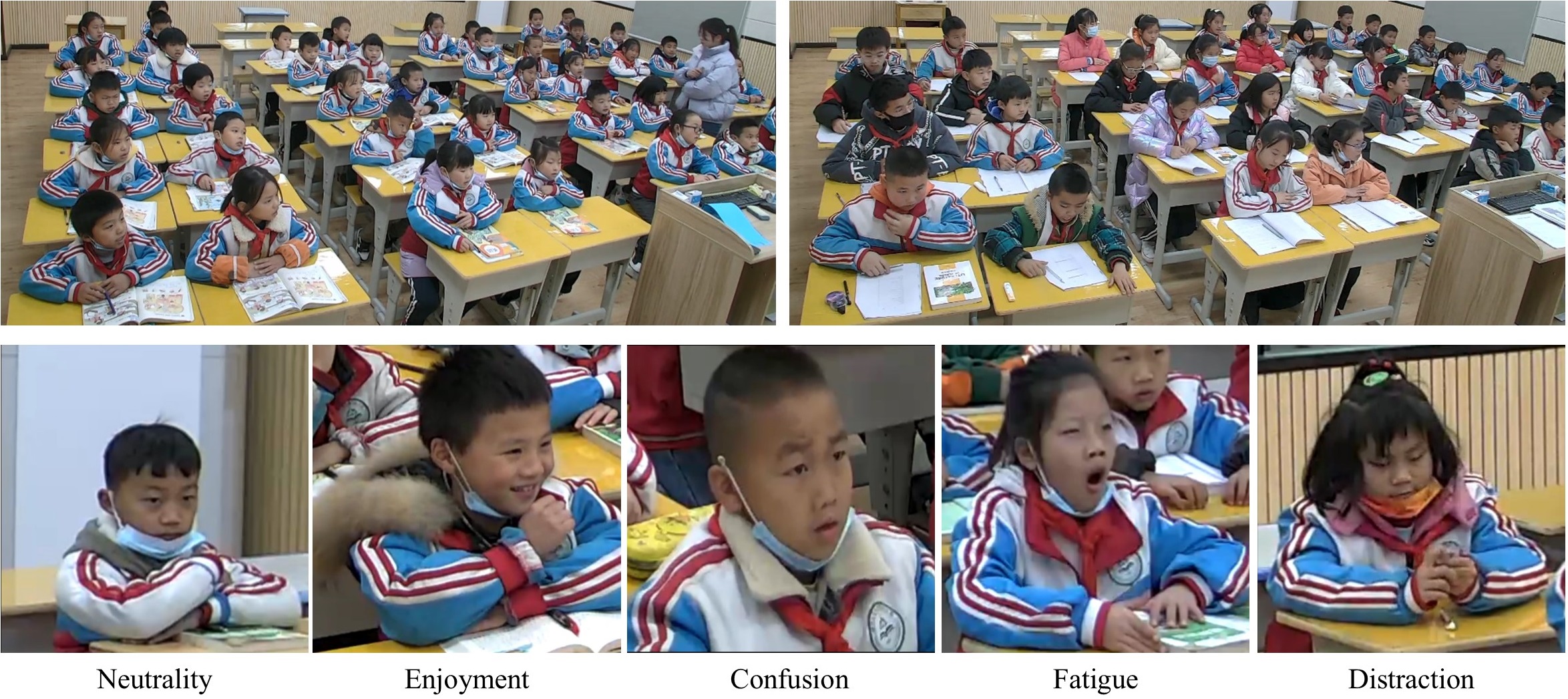}
    \vspace{-5mm}
    \caption{Examples from the JuniorRAER dataset. Top row: original video frames; Bottom row: samples of the 5-class academic emotions.
 }
    \label{fig:junior_dataset}
    \vspace{-5mm}
\end{figure*}

\section*{B. Additional Implementation Details}
\noindent{\textbf{Website for Academic Emotion Annotation.}} Fig.~\ref{fig:website} presents the user interface (UI) of the annotation website developed for labeling academic emotion videos. Through this UI, each annotator can individually review the video clips assigned to them and select an emotion category from the given coarse-grained or fine-grained academic emotion label sets. For challenging videos, annotators can use the progress bar to repeatedly view the video to determine the most appropriate emotion category.

\noindent\textbf{Dataset of JuniorRAER.} We built a small academic emotion dataset, called JuniorRAER, to evaluate the generalization ability of our model. Specifically, video recordings in classrooms are common in first-tier cities, generating large amounts of data daily. We sourced the original video data from 6 open video-recorded courses, with the consent of both teachers and students for research analysis. This dataset captures the emotions of primary school students, approximately 10 years old, in real classroom settings. We processed and annotated these videos using methods similar to those described in Sec.~\ref{sec:collect_video} and Sec.~\ref{sec:label_video} of the main document. As a result, we obtained 468 academic emotion video clips featuring 35 primary school students (17 male and 18 female). Similar to the RAER dataset, the JuniorRAER dataset also exhibits an imbalanced distribution: 351 clips (75\%) are labeled as neutrality, 61 clips (13.03\%) as distraction, 28 clips (5.98\%) as fatigue, 24 clips (5.13\%) as enjoyment, and 4 clips (0.85\%) as confusion. We split the JuniorRAER dataset into training (60\%) and testing (40\%) sets, ensuring a nearly identical distribution of academic emotions in both subsets. To protect privacy, this dataset is strictly for non-commercial research purposes.
Fig~\ref{fig:junior_dataset} illustrates the original videos alongside specific examples of the 5-class academic emotions.

\noindent{\textbf{Descriptors of Academic Emotion Categories.}}  
In the context-aware text encoder of the proposed CLIP-CAER, we employ a large language model, such as ChatGPT~\cite{ChatGPT-2018}, to generate text descriptions for each academic emotion category, capturing both the associated facial expressions and relevant context behaviors. To generate descriptors, we first provide the LLM with the classification criteria outlined in Sec.~\ref{sec:label_video} of the main document, allowing it to form a memory. Then, we use the prompt: ``What are the useful visual features for the academic emotion of \{classname\}, considering both facial characteristics and context information?'' The descriptors for each emotion category are as follows:
\begin{itemize}
    \item \textbf{Neutrality:} Relaxed mouth, open eyes, neutral eyebrows, no noticeable emotional changes, engaged with study materials, or natural body posture.
    \item  \textbf{Enjoyment:} Upturned mouth corners, sparkling eyes, relaxed eyebrows, focused on course content, or occasionally nodding in agreement.
    \item \textbf{Confusion:} Furrowed eyebrows, slightly open mouth, wandering or puzzled gaze, chin rests on the palm, or eyes lock on learning material.
    \item \textbf{Fatigue:} Mouth opens in a yawn, eyelids droop, head tilts forward, eyes lock on learning material, or hand writing.
    \item \textbf{Distraction:} Shifting eyes, restless or fidgety posture, relaxed but unfocused expression, frequently checking phone, or averted gaze from study materials.
\end{itemize}
Note that each descriptor consists of two parts: the first describes the corresponding facial expression behaviors, while the second captures relevant contextual learning behaviors, such as body posture, yawning, or using a phone. If only facial expression information from the video sequence is used, the text descriptions are modified to include only the facial expression behavior component.

\noindent\textbf{Optimization.}
We train the entire network using the SGD optimizer with a batch size of 8. The base learning rates are set as follows: \(1 \times 10^{-5}\) for the CLIP image encoder, \(1 \times 10^{-2}\) for the temporal visual encoder, \(1 \times 10^{-3}\) for the learnable prompt, and \(1 \times 10^{-5}\) for the fully connected layer. The learning rates are reduced by an order of magnitude at the 10\textsuperscript{th} and 15\textsuperscript{th} epochs. The model is trained for 20 epochs in an end-to-end manner.
For each video, we randomly and uniformly select 16 non-overlapping frames and use a face detector to extract the face region from each frame, both of which are used as inputs to the model. Each full frame or face region is resized to \(224 \times 224\), with the shorter edge padded in black to match the input size required by the CLIP model~\cite{CLIP-ICML2021}. During training, we apply data augmentation techniques, including random rotation and random flipping, to enhance robustness.

\section*{C. Experiments on CAER Dataset}
Table~\ref{tab:comparison_CAER} presents the evaluation of our method on the CAER dataset~\cite{ContextFER-ICCV2019}, which focuses on basic emotions. The CAER dataset includes not only facial expressions but also rich context information, providing a comprehensive representation of emotional responses. However, unlike the academic emotion dataset RAER, the CAER dataset focuses on basic emotions and is substantially larger, with over 13,200 videos. In our implementation, for the seven basic emotions in CAER, we used descriptors similar to those in ~\cite{DFER-CLIP2023} to describe facial expressions while also providing additional context descriptions. The details are as follows:
\begin{itemize}
    \item \textbf{Surprise:} Widened eyes, an open mouth, raised eyebrows, and a frozen expression. Sudden stillness, widened eyes on the other person, hands raised or paused mid-motion.
\item \textbf{Sad:} Tears, a downward-turned mouth, drooping upper eyelids, and a wrinkled forehead. Head down, avoiding eye contact, slow, withdrawn movements.
\item \textbf{Neutral:} Relaxed facial muscles, a straight mouth, a smooth forehead, and unremarkable eyebrows. Relaxed posture, open stance, steady, calm eye contact.
\item \textbf{Happy:} A smiling mouth, raised cheeks, wrinkled eyes, and arched eyebrows. Leaning in toward the other person, quick, cheerful movements.
\item \textbf{Fear:} Raised eyebrows, parted lips, a furrowed brow, and a retracted chin. Hands close to chest or tightly together, small, cautious steps backward.
\item \textbf{Disgust:} A wrinkled nose, lowered eyebrows, a tightened mouth, and narrow eyes. Slight step back, body angled away, hand raised or shielding face.
\item \textbf{Anger:} Furrowed eyebrows, narrow eyes, tightened lips, and flared nostrils. Leaning forward, tense stance, fists clenched, or hand pointing.
\end{itemize}
It can be observed from Table~\ref{tab:comparison_CAER} that incorporating context information in addition to facial expressions significantly improves recognition performance. Furthermore, compared to the state-of-the-art method CAER-Net~\cite{ContextFER-ICCV2019}, which also leverages context, the proposed CLIP-CAER achieves an improvement of up to \textbf{4.73} points, reaching an accuracy of \textbf{81.77\%}. These results demonstrate that the proposed CLIP-CAER method is highly effective for both academic and basic emotion recognition, significantly surpassing current state-of-the-art methods.

\begin{table}
    \centering
        \caption{Evaluation of CLIP-CAER compared to 3DCNN~\cite{3DCNN-CVPR2018} and CAER-Net~\cite{ContextFER-ICCV2019} on the CAER benchmark for basic emotions.}
        \vspace{-2mm}
    \label{tab:comparison_CAER}
    \begin{tabular}{lc}
    \hline
       Method &  UAR(\%) \\
        \hline
        \hline
        3DResNets18~\cite{3DCNN-CVPR2018} w/o Context & 68.22\\ 
        {CAER-Net~\cite{ContextFER-ICCV2019}} w/o Context \quad\quad\quad\quad\quad\quad\quad & 74.13 \\
         {CAER-Net~\cite{ContextFER-ICCV2019}} w/ Context & 77.04\\
         \hline
        {CLIP-CAER} w/o Context & 75.36 \\
         {CLIP-CAER} w/ Context &  \textbf{81.77} \\
         \hline
    \end{tabular}
\end{table}

\begin{table}
    \centering
    \caption{Ablation study on the number of layers in the temporal encoder and learnable prompt tokens. '\# Layer' and '\# Tokens' denote the number of layers in the temporal encoder and learnable prompt tokens, respectively.}
    \vspace{-2mm}
    \label{tab:ablation_number}
    \setlength{\tabcolsep}{4mm}{
    \begin{tabular}{c|c|c}
        \hline
      \# Layer & \# Tokens & UAR(\%) \\ 
       \hline
       \hline
       1 & 8 & \textbf{68.00\%} \\
       2 & 8 & 64.78\% \\
       3 & 8 & 64.29\% \\
         \hline
       1 & 4	& 65.54\% \\
       1 & 8	& \textbf{68.00\%} \\
       1 & 12	& 64.15\% \\
       1 & 16	& 64.27\% \\
         \hline
    \end{tabular}
    }
\end{table}

\begin{figure*}
    \centering
    \includegraphics[width=1.0\linewidth]{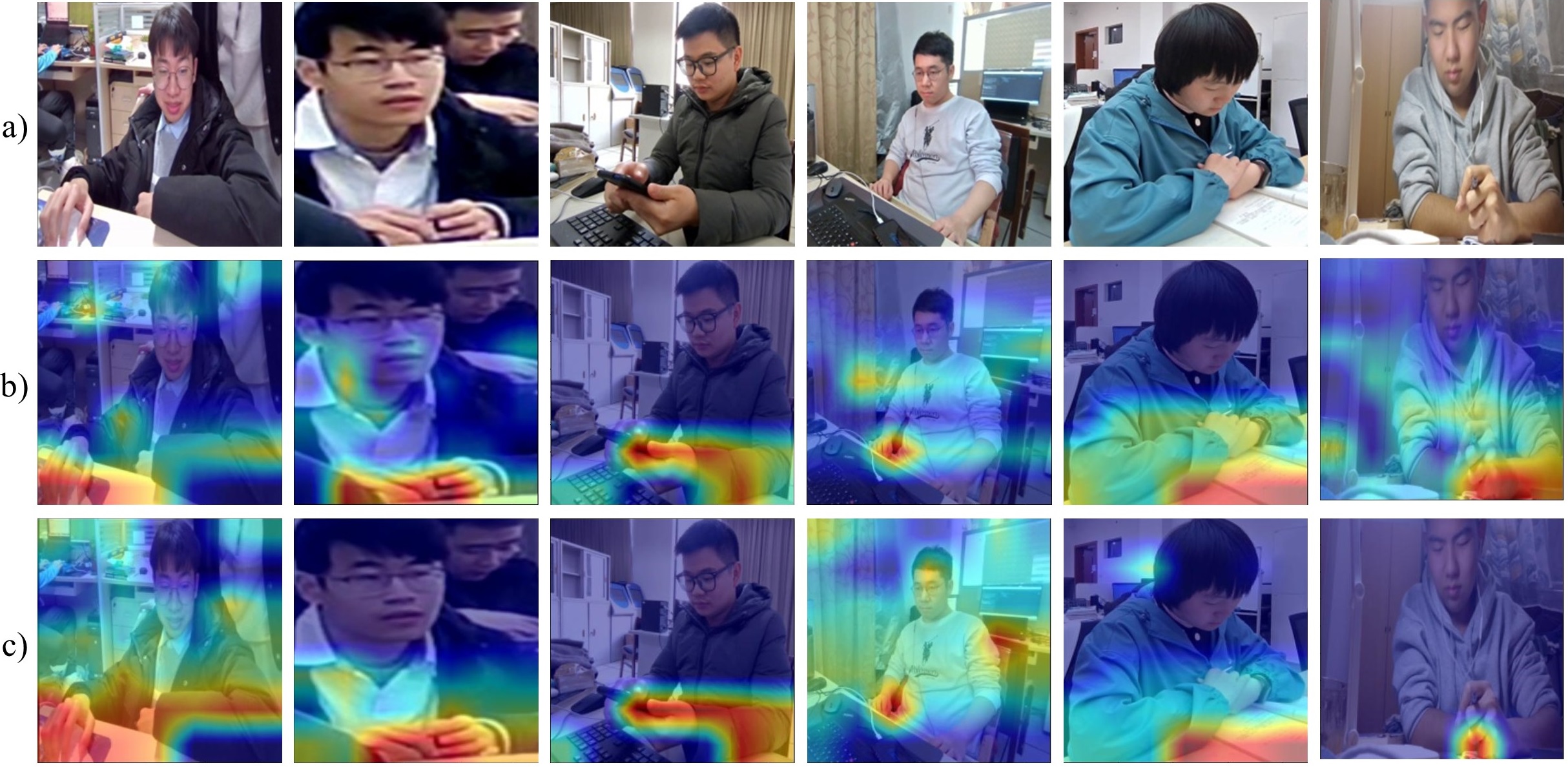}
    \caption{Visualization of attention on full-frame images using Grad-CAM~\cite{GradCAM-ICCV2017}: a) Full-frame input; b) Model utilizing full-frame video sequences to jointly capture facial expressions and context information; c) Model integrating facial image sequences with full-frame sequences to separately capture facial expressions and context information.}
    \label{fig:vis_compare}
\end{figure*}

\section*{D. Additional Ablation Study}
\textbf{Number of Layers in the Temporal Encoder and Learnable Prompt Tokens. } 
Table~\ref{tab:ablation_number} examines the impact of varying the number of layers in the temporal encoder and the effect of different numbers of learnable prompt tokens for each category. The self-attention module \(\text{S-ATT}\) in Eq.~\ref{eqn:self_attention}, used in the context-aware temporal visual encoder, consists of several identical self-attention layers sequentially stacked together~\cite{Attention_NIPS2017}. In general, increasing the number of layers tends to improve model performance; however, in the proposed CLIP-CAER, the best performance is achieved with a single-layer temporal encoder. This finding aligns with the conclusion in~\cite{DFER-CLIP2023}. The primary reason is that the academic emotion dataset RAER is relatively small, and the temporal encoder is trained from scratch, which makes it prone to overfitting if the model is overly complex, thereby degrading generalization performance on the test data. This consideration also applies to the learnable prompt tokens, where using 8 learnable tokens achieves the best performance, while increasing the number of tokens does not lead to further improvements. 

\noindent\textbf{Attention Visualization.}
Fig.~\ref{fig:vis_compare} visualizes the attention regions of our model on input images using Grad-CAM~\cite{GradCAM-ICCV2017}. In this visualization, we compare two input strategies for the model: (a) using full-frame video sequences to jointly capture facial expression features and context information, and (b) combining facial image sequences with full-frame sequences to separately model facial expression features and context information.
As shown, both strategies effectively capture context information relevant to academic emotion recognition within the full-frame images, such as using a phone or reading a book,  thanks to the robust alignment between text and visual feature spaces provided by the pre-trained CLIP model.
However, both tend to overlook facial expression information in the full frame due to the relatively small size of the face region compared to the surrounding context, which may cause the model to disregard it.
To address this issue, our model adopts strategy (b), which incorporates an additional facial image sequence to specifically capture facial expression features. These results highlight the effectiveness and robustness of our model's design, as well as the importance of incorporating both facial image sequences and full-frame sequences as inputs.

\section*{E. Ethical Implications} 

The RAER dataset, which consists of real-world academic emotion videos, involves the collection and processing of student data. Ensuring privacy protection is paramount, given that facial expressions and context cues are sensitive personal data. To address this, we adhere to strict data anonymization protocols, ensuring that personally identifiable information is removed. 
Storage and access control mechanisms are also implemented to prevent unauthorized use of the dataset. Data sharing is regulated to ensure compliance with relevant legal and ethical standards, such as local data protection laws. Researchers using RAER must agree to ethical data usage policies to minimize the risk of privacy breaches.

To mitigate potential bias across different cultural backgrounds, we introduce JuniorRAER as an indirect validation of the model's generalization ability. However, this does not fully resolve cross-cultural bias, as learning environments and emotional expressions can vary significantly across cultures. Future work should focus on diversifying the dataset by including students from different ethnicities, socioeconomic backgrounds, and educational settings to enhance fairness and robustness. Additionally, the subjective nature of emotion annotation introduces another layer of bias. Human annotators may interpret emotions differently based on their own experiences and cultural backgrounds, leading to inconsistencies in labeling. To reduce this bias, we employed a majority voting strategy across multiple annotators, ensuring greater reliability in emotion classification. Further studies could explore leveraging self-reported emotions or multimodal signals to enhance label accuracy.

The real-world application of academic emotion recognition systems must be approached with caution. While such models have the potential to enhance personalized learning and provide insights into student engagement, they also carry risks of misuse. For instance, over-reliance on AI-based emotion recognition in educational settings could lead to unintended consequences, such as automated decision-making that lacks human oversight.
To prevent ethical misuse, AI-based academic emotion recognition systems should be used as assistive tools rather than absolute evaluators of student emotions or performance. Educators and stakeholders must be trained to interpret model outputs critically, using them to complement rather than replace human judgment. 

\end{document}